\documentclass[preprint]{elsarticle}

\usepackage{lineno,hyperref}
\modulolinenumbers[5]

\usepackage{amsmath}
\usepackage{url}

\biboptions{authoryear}

\begin{document}

\begin{frontmatter}

\title{Ranking and significance of variable-length similarity-based time series motifs}

\author[iiia]{Joan Serr\`a}\ead{jserra@iiia.csic.es}
\author[crm]{Isabel Serra}\ead{iserra@crm.cat}
\author[crm]{\'Alvaro Corral}\ead{acorral@crm.cat}
\author[iiia]{Josep Lluis Arcos}\ead{arcos@iiia.csic.es}

\address[iiia]{Artificial Intelligence Research Institute (IIIA-CSIC), Bellaterra, Barcelona, Spain}
\address[crm]{Centre de Recerca Matem\`atica, Bellaterra, Barcelona, Spain}

\begin{abstract}
The detection of very similar patterns in a time series, commonly called motifs, has received continuous and increasing attention from diverse scientific communities. In particular, recent approaches for discovering similar motifs of different lengths have been proposed. In this work, we show that such variable-length similarity-based motifs cannot be directly compared, and hence ranked, by their normalized dissimilarities. Specifically, we find that length-normalized motif dissimilarities still have intrinsic dependencies on the motif length, and that lowest dissimilarities are particularly affected by this dependency. Moreover, we find that such dependencies are generally non-linear and change with the considered data set and dissimilarity measure. Based on these findings, we propose a solution to rank those motifs and measure their significance. This solution relies on a compact but accurate model of the dissimilarity space, using a beta distribution with three parameters that depend on the motif length in a non-linear way. We believe the incomparability of variable-length dissimilarities could go beyond the field of time series, and that similar modeling strategies as the one used here could be of help in a more broad context.
\end{abstract}

\begin{keyword}
Time series \sep Motif ranking \sep Distance modeling \sep Beta distribution
\end{keyword}

\end{frontmatter}


\section{Introduction}
\label{sec:intro}

Time series typically represent a record of the underlying dynamics of a process or system. When appropriate measurements are taken, the information contained in a time series can be crucial to understand and/or model such a system. In particular, the detection of repeated or very similar patterns in a time series, commonly called motifs, has shown to be of great value for researchers and practitioners. Examples range from natural and health sciences to animation or business analytics~\citep{Mueen14WIRES}.

In general, two formal definitions of a time series motif coexist. The first one is based on the notion of frequency~\citep{Lin02WTDM}: a pattern is interesting if it has a significant amount of repetitions. The second one is based on the notion of similarity~\citep{Mueen09SDM}: a pattern is interesting if its occurrences are identical or too similar to happen at random. Both definitions are complementary, as a strikingly similar pattern does not necessarily need to be frequent. Hence, algorithms exploiting both notions independently have received continuous and increasing attention~\citep{Chiu03KDD,Tanaka05ML,Mueen09SDM,Tang08KBS,Rakthanmanon11ICDM,Mueen13ICDM,Yingcharxxx13ICDM}.

Under a frequency-based definition, the ranking of the motifs found in a time series is trivial. The most important motif is the one with the highest count, the second most important motif is the one with the second highest count, and so on. Moreover, we can assess the statistical significance of frequency-based motifs by comparing observed and expected counts under a null model reflecting some basic characteristics of the time series. This has been exploited by \citet{Castro11SDM}, who leverage work from the bioinformatics community to derive a motif's significance.

Using a similarity-based definition, motif ranking also looks straightforward. Given a single (usually pre-specified) motif length, the most important motif pair is the one with the lowest dissimilarity, the second most important pair is the one with the second lowest dissimilarity, and so on (equivalently for highest similarity). However, if we have motif pairs of different lengths, we cannot directly compare dissimilarities or distances, as these typically depend on the length of the given segments. In these cases, researchers rely on two different strategies. On the one hand, there is the option to compute a ranking for every motif length of interest, possibly removing covering motifs~\citep[e.g.,][]{Mueen13ICDM}. Consequently, we have as many orderings as lengths being considered, and the choice for the most important motif depends on the user. On the other hand, there is the possibility to normalize the dissimilarity measure by the length of the motif, or to use a measure that already incorporates some notion of normalization\footnote{All dissimilarities considered in this paper are normalized by the length of the motif (sometimes we will additionally employ the terms ``normalized'' or ``length-normalized'' to further clarify this aspect). The reader should not confuse these terms with the typical z-normalization between time series or other possible normalization strategies (see also Sec.~\ref{sec:mm_dissim}).}~\cite[e.g.,][]{Yingcharxxx13ICDM}. For instance, one can divide the Euclidean distance by the square root of the length, or consider the Pearson's correlation measure. In terms of motif significance, similarity-based approaches are much less developed than frequency-based ones. In fact, to the best of our knowledge, this topic has not been considered yet.

In this work, we show an important and overlooked aspect of variable-length similarity-based motifs: that they cannot be directly compared, and hence ranked, using common motif dissimilarity measures and their corresponding length normalization. Using a variety of statistical tools, we illustrate that normalized motif dissimilarities exhibit intrinsic dependencies with respect to the motif length, and that these particularly affect the lowest dissimilarities of each length. Moreover, we find that such dependencies are generally non-linear, and change with the considered measure and data set. These aspects are quantified using a combination of 8~common dissimilarity measures and 9~different publicly-available time series data sets. To further facilitate the assessment and reproducibility of our work, we make all results and code available online.

Given the aforementioned problems, and as a further contribution, we propose a solution to compare motifs of different lengths and, at the same time, derive a measure of their significance. The proposed solution consists of a compact model of the motif dissimilarity space, using a beta distribution whose parameters non-linearly depend on the length of the motif. We find this model to yield a reasonable fit for the majority of the considered lengths, measures, and data sets. Importantly, the cumulative distribution function (CDF) of the proposed model can wrap the motif dissimilarity function, hence directly yielding a $p$-value for each motif pair that can be used for ranking and significance assessment inside the motif discovery algorithm.

The remainder of the article is structured as follows. Sec.~\ref{sec:comparison} analyzes the problem of comparing motifs of different lengths. Sec.~\ref{sec:model} introduces the proposed modeling strategy. Sec.~\ref{sec:mm} gives the details of the considered materials and methods. Sec.~\ref{sec:conclusion} concludes the article by summarizing our work and highlighting some future directions.

\section{Comparing motif dissimilarities}
\label{sec:comparison}

\subsection{Motivating examples}
\label{sec:comparison_examples}

\begin{figure}[t]
	\centering
	\includegraphics{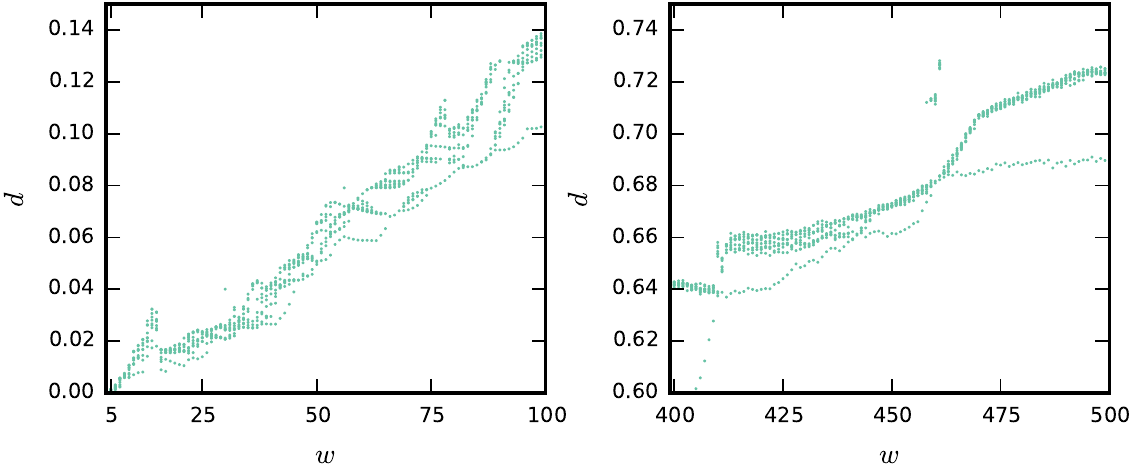}
	\caption{Lowest 10 dissimilarities $d$ for each segment length $w$ considering all possible non-overlapping segments from a section of the \textsc{EEG} data set: $w\in[5,100]$ (left) and $w\in[400,500]$ (right). Importantly, notice that length-normalized Euclidean distance is used (see main text and also Sec.~\ref{sec:mm_dissim}).}
	\label{fig:example_moen}
\end{figure}

To understand the issues that arise when comparing variable-length motif dissimilarities, we first take a look at some examples. Let's consider a randomly-chosen contiguous segment of 10,000 samples from the \textsc{EEG} data set (Sec.~\ref{sec:mm_data}). We then compute the length-normalized Euclidean distance $d$ (Sec.~\ref{sec:mm_dissim}) between all possible non-overlapping pairs of segments of length $w\in[5,100]$, and take the lowest 10 dissimilarities $d$ for each $w$. What we observe is a clear trend of increasing $d$ with $w$ (Fig.~\ref{fig:example_moen}, left). Given this trend, how can we automatically determine the most important motif using a similarity-based definition? To make it more explicit, let's assume that the best motif at length $w_1=30$ scores a length-normalized distance of $d_1=0.202$ and that the best motif at length $w_2=40$ scores a length-normalized distance $d_2=0.219$. Based on what we have seen (Fig.~\ref{fig:example_moen}, left), which one should we prefer? Notice that, furthermore, both motifs could overlap. Would we prefer motif 2, an extension of motif 1, even if the length-normalized dissimilarity is not as low as the one of motif 1? How can we choose in an objective and informed way? This are the kinds of situations this work deals with. However, we first need to demonstrate that such situation is systematically occurring, independently of the data source and the dissimilarity measurement.

Re-taking our motivating example (Fig.~\ref{fig:example_moen}), we could argue that the observed trend is due to the short length of the segments ($w\in[5,100]$). However, if we repeat the calculations for $w\in[400,500]$, another trend appears (Fig.~\ref{fig:example_moen}, right). Notice the change in the dissimilarity values, which is more than 4 times larger (Fig.~\ref{fig:example_moen}, vertical axes). Such a difference is difficult to attribute to the effect of some characteristic timescales. Instead, it looks as a property of both the combination time series and the used dissimilarity measure (see below).

\begin{figure}[t]
	\centering
	\includegraphics{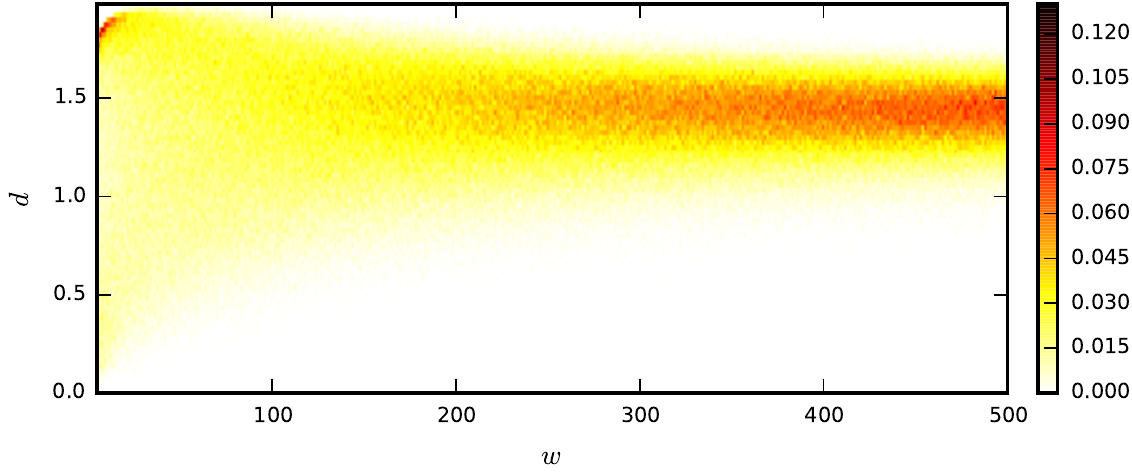}
	\caption{Normalized histogram of length-normalized Euclidean distances using $n=2000$ dissimilarity samples for each $w\in[5,500]$ and the full \textsc{EEG} data set.}
	\label{fig:example_hist}
\end{figure}

The aforementioned trends are clearly observable by a simple uniform random sampling of the motif dissimilarity space. If, for each $w\in[w_{\text{min}},w_{\text{max}}]$, we select $n$ non-overlapping segment pairs at random and compute their length-normalized Euclidean distance, we can reproduce the same phenomenon (Fig.~\ref{fig:example_hist}). The plotted histogram gives us an indication that the empirical distribution of $d$ changes with $w$. As $w$ increases, the mode of the distribution seems to be more or less stable, but the tails (i.e.,~the non-central parts of the distribution) are visibly different, specially the lower one.

\begin{figure}[t]
	\centering
	\includegraphics{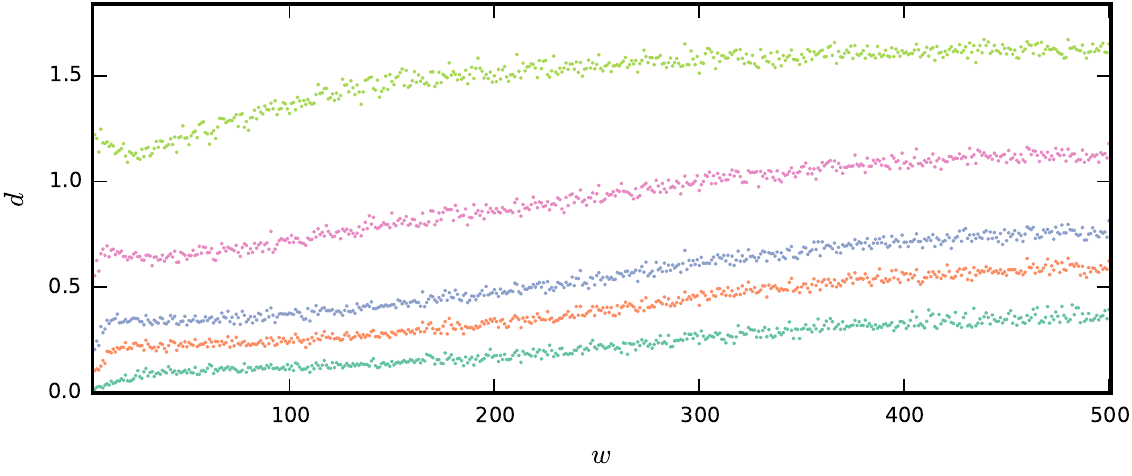}
	\caption{Quantiles for a sample of length-normalized dissimilarities using dynamic time warping (DTW) and the \textsc{Wind} data set. From top to bottom, the quantiles correspond to 0.5, 0.25, 0.1, 0.05, and 0.01.}
	\label{fig:example_quants}
\end{figure}

With further analysis, we confirm that the observed phenomenon is not unique of the \textsc{EEG} data set nor of the normalized Euclidean distance. In fact, if we consider other data sources and dissimilarities with their corresponding length normalization (Sec.~\ref{sec:mm_dissim}), we can easily obtain more radical examples of the same phenomenon (see, e.g., Fig.~\ref{fig:example_quants}). In this example, we can compute the statistical significance of the slopes of the plotted quantiles, obtained via ordinary least squares, for $w\in[300,500]$. The highest $p$-value we obtain is $p=1.93\cdot 10^{-15}$, which corresponds to the slope of the median. Thus, we see that even the median can show a statistically significant trend for relatively large $w$. The plotted example also depicts a non-linear dependency of the computed quantiles with respect to $w$ (Fig.~\ref{fig:example_quants}). We can also observe that such dependency is different than the one seen in Fig.~\ref{fig:example_hist}. Clear differences are observable even if we fix the data source and change the dissimilarity measurement. This suggests that the observed behaviors are not due, to a large extent, to the effect of some characteristic timescales of the time series.

\subsection{Quantitative evaluation}
\label{sec:comparison_formal}

To quantify the incomparability of $d$ with respect to $w$ in a more formal and rigorous way, we employ two basic measures of the difference between distributions. First, we consider the global disagreement between empirical CDFs. We quantify it using
\begin{equation*}
	\epsilon = \frac{1}{k} \sum_{i=1}^{k} \left| F_i - G_i \right| ,
\end{equation*}
where $k$ is an arbitrarily chosen bin resolution and $F$ and $G$ correspond to the two empirical CDFs being compared. Notice that $\epsilon$ is conceptually similar to the total variation distance between probability distributions~\citep{Levin09BOOK}. Nonetheless, since we use CDFs and take the average, $\epsilon\in[0,1]$ gives a rapid and intuitive idea of the average difference between distributions. The bin resolution for all experiments reported here corresponds to $k=100$ equally-spaced bins between the minimum and the maximum of each sample for each $w$.

As we are interested in the best motifs, we need to pay special attention to the lower tails of the dissimilarity distributions (i.e.,~the lowest sample values for each $w$). Hence, we consider a second measure based on just the lowest quartile of the empirical sample. Specifically, we consider the well-known Kolmogorov-Smirnov (KS) test~\citep{Massey51JASA} and its associated $p$-value, which we denote by $p_{\text{KS}}$. The KS test is a non-parametric test of the equality of continuous, one-dimensional probability distributions that can be used to compare a sample with a reference probability distribution or to compare two samples. In our case, we compare the first quartile of the two samples.

\begin{figure}[t]
	\centering
	\includegraphics{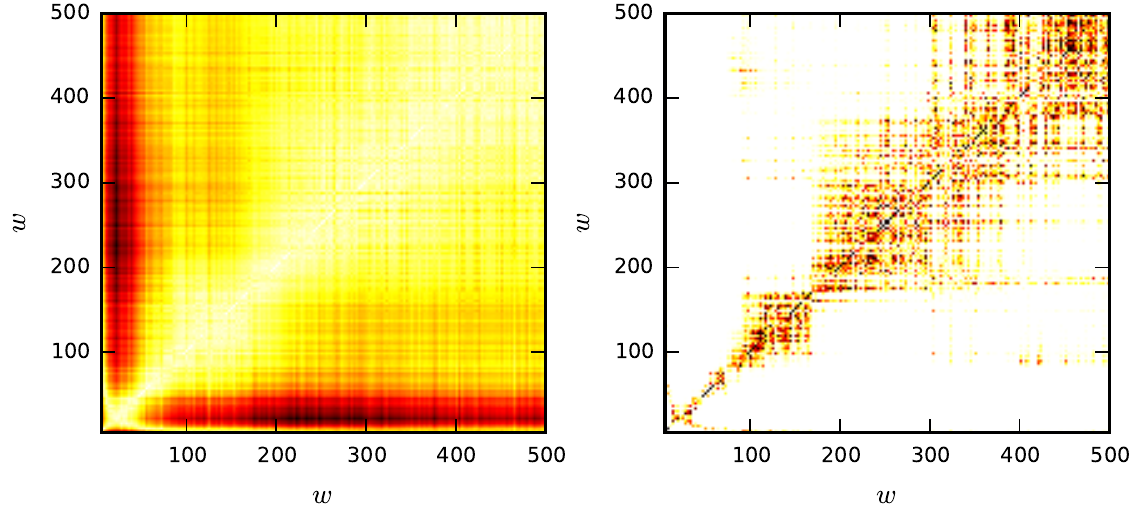}
	\caption{Matrices comparing distribution differences between all possible pairwise comparisons in $w$: $\epsilon$ (left) and $p_{\text{KS}}$ (right). The samples come from using correlation and the \textsc{CarCount} data set. The color code goes from 0 (white) to 0.1 and 1 (black), respectively.}
	\label{fig:example_epspval}
\end{figure}

\begin{figure}[t]
	\centering
	\includegraphics{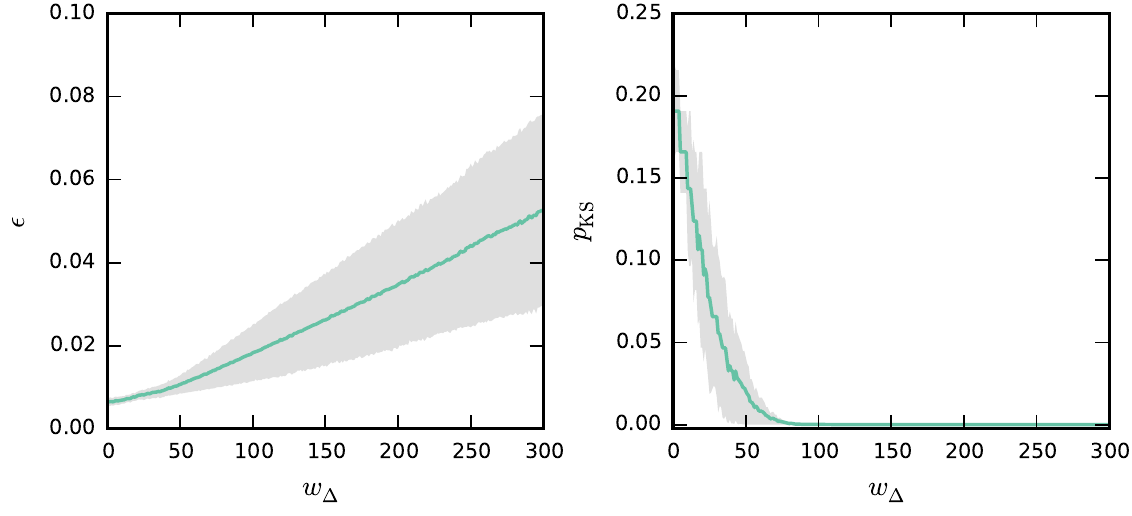}
	\caption{Median (line) and median absolute deviation (patches) for $\epsilon$ (left) and $p_{\text{KS}}$ (right), displayed as a function of $w_\Delta$. Results computed from aggregating all data sets and measures (see text).}
	\label{fig:w_epspval}
\end{figure}

Computing $\epsilon$ and $p_{\text{KS}}$ for all possible pairwise comparisons of samples in $w$ yields two matrices that can be post-processed in order to aggregate the information for each data set and dissimilarity measure (Fig.~\ref{fig:example_epspval}). If, for a given data set and measure, we take statistics of the diagonals of these matrices, we obtain an assessment of the distribution differences as a function of $w_\Delta=|w_i-w_j|$, the absolute difference between two motif lengths $w_i$ and $w_j$. Specifically, for a given $w_\Delta$, we compute the median and the median absolute deviation of $\epsilon$ and $p_{\text{KS}}$. Aggregating these results for all possible combinations of data set and dissimilarity measure gives us an idea of the expected differences when comparing two distributions separated by $w_\Delta$ (Fig.~\ref{fig:w_epspval}). For instance, if we compare a motif pair of length $w_i=150$ with a motif pair of length $w_j=190$ ($w_\Delta=40$), we can expect an average CDF error $\epsilon\approx 0.01$ and a $p_{\text{KS}}\approx 0.03$ (Fig.~\ref{fig:w_epspval}). The former tells us that, on average, there will be a difference between CDFs of one per cent. The latter tells us that the tails of the distributions are hardly comparable, given that $p_{\text{KS}}$ is systematically lower than the significance threshold of 0.05. Thus, in general, we see that comparing motifs whose lengths differ by more than 40 samples is hardly justifiable. Further details can be found in the online results (Sec.~\ref{sec:mm_download}).

\begin{figure}[t]
	\centering
	\includegraphics{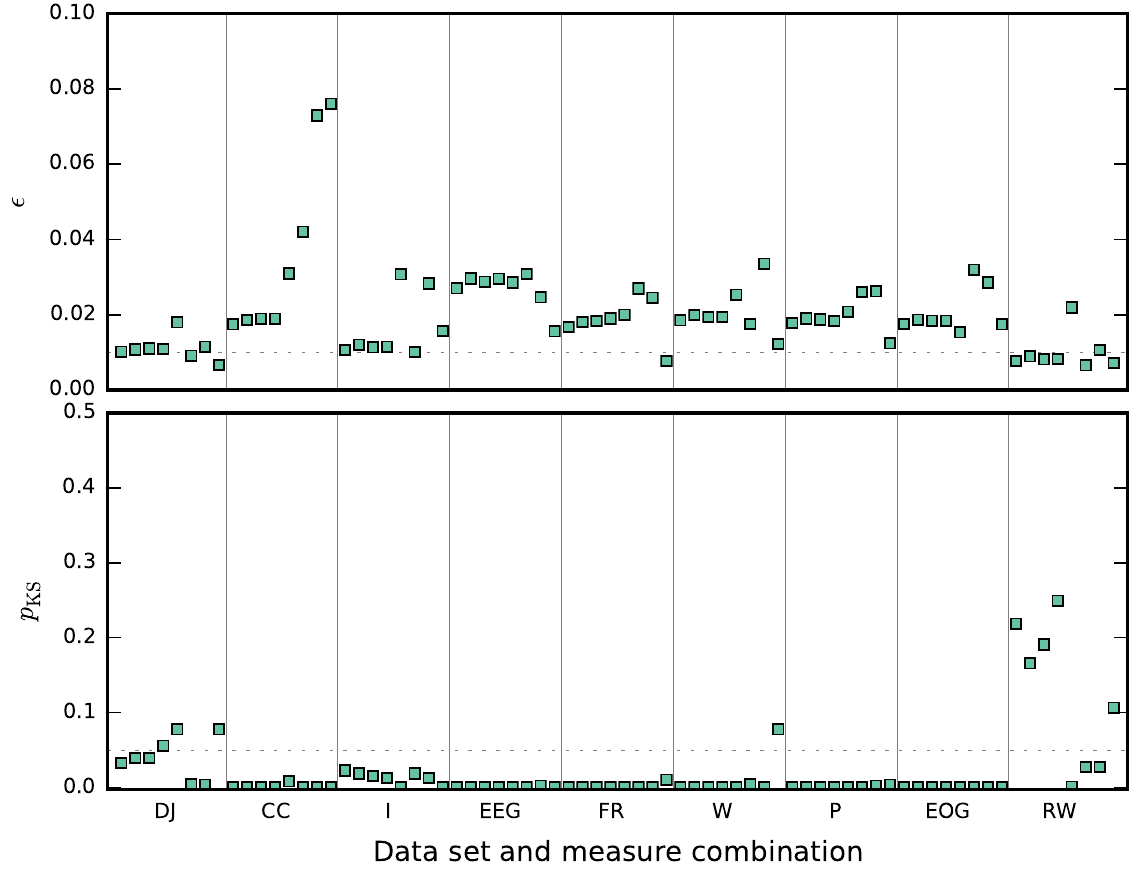}
	\caption{Incomparability of distributions: median values for $\epsilon$ (top) and $p_{\text{KS}}$ (bottom) for $w_\Delta=100$ and every studied combination of data set and dissimilarity measure. There are a total of $9\times 8=72$ such combinations (Secs.~\ref{sec:mm_data} and~\ref{sec:mm_dissim}). Vertical lines separate data set blocks: \textsc{DowJones} (DJ), \textsc{CarCount} (CC), \textsc{Insect} (I), \textsc{EEG} (EEG), \textsc{FieldRecording} (FR), \textsc{Wind} (W), \textsc{Power} (P), \textsc{EOG} (EOG), and \textsc{RandomWalk} (RW).}
	\label{fig:w_epspvalthres}
\end{figure}

Let's take $w_\Delta=100$ and analyze the results for individual combinations of data set and measure (Fig.~\ref{fig:w_epspvalthres}). We observe that nearly all distribution differences $\epsilon$ are above 0.01 and that almost no combination passes the KS test at a significance level of $p_{\text{KS}}>0.05$. However, there is one notable exception: the last 8 combinations, which correspond to the \textsc{RandomWalk} data (Fig.~\ref{fig:w_epspvalthres}, right). Several dissimilarity measures on this data set achieve acceptable $p_{\text{KS}}$ values while keeping $\epsilon\approx 0.01$. This is to be expected, and tells us that, for the case of artificially generated random Gaussian data (Sec.~\ref{sec:mm_data}), length-normalized motif dissimilarities tend to comparable, even across very different lengths. Apart from the \textsc{RandomWalk} data, the first 8 combinations, which correspond to the \textsc{DowJones} data, seem to achieve larger $p_{\text{KS}}$ values than the rest (Fig.~\ref{fig:w_epspvalthres}, left). This is interesting, as in economics the random walk hypothesis has been used to model share prices and other factors for a long time~\citep{Malkiel73BOOK}.

\section{Modeling motif dissimilarities}
\label{sec:model}

\subsection{Main idea}
\label{sec:model_idea}

To overcome the drawbacks described in the previous section, we now propose a procedure to model the dissimilarity space. Our aim is to produce a compact model of the empirical dissimilarity distributions for each $w$ from a given combination of data set and dissimilarity measure. The main idea behind our modeling strategy is to achieve a `normalization' of the dissimilarity space. We want to transform the dissimilarity space into a uniform probability space in which motifs of different lengths can be compared in a meaningful way. 

In Sec.~\ref{sec:comparison}, we have seen that, given two length-normalized dissimilarities $d_i$ and $d_j$ obtained from $w_i$ and $w_j$, respectively, the relation $d_i<d_j$ does not necessarily imply that $d_j$ should be ranked after $d_i$. Our observation is that, by considering an estimated CDF for each $w$, we can mix motifs of different lengths and meaningfully compare them. For instance, if $P_w(D\leq d)$ denotes the estimated CDF of the dissimilarities for a fixed $w$, then $P_{w_i}(D\leq d_i) < P_{w_j}(D\leq d_j)$ implies that $d_j$ should indeed be ranked after $d_i$. We will further develop this idea, and specially the way to estimate $P_w$, in the next sections. In the end, we plan to substitute a given dissimilarity $d$ by $d'=P_w(D\leq d)$.

\subsection{Preliminary analysis}
\label{sec:model_preliminary}

\begin{figure}[t]
	\centering
	\includegraphics{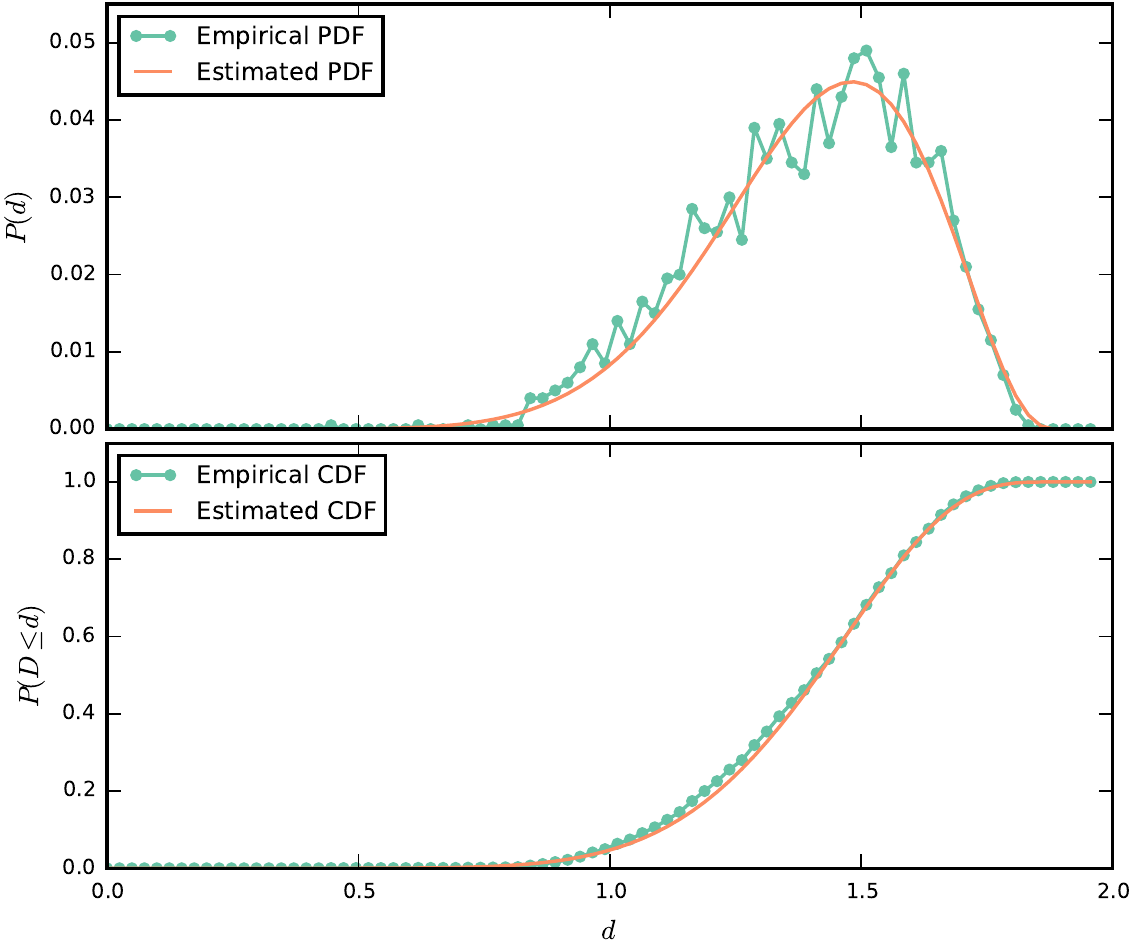}
	\caption{Examples of an empirical PDF (top) and CDF (bottom) and their estimated fits. The sample comes from taking the length-normalized Euclidean distance and the \textsc{Insect} data set with $w=460$. Here, our procedure estimates $\alpha=9.37$, $\beta=3.03$, and $m=1.93$.}
	\label{fig:fit_pdfcdf}
\end{figure}

An illustration of the empirical probability distribution function (PDF) for dissimilarities with fixed $w$ is shown in Fig.~\ref{fig:fit_pdfcdf}. Observe that a Gaussian model could initially appear as a reasonable model. However, this is not so. The Gaussian model is a good model for the central part of an empirical distribution, but it has the limitation that the kurtosis is always zero. Hence, it does not correctly model the observed tails. Contrastingly, the similarity-based motif discovery task requires to get accurate estimations at the tails of such distributions. In fact, we are only interested in the smallest existing dissimilarities. Thus, our modeling task requires a good model for the tails. In particular, it requires a model with a good fit in the left, lowest dissimilarity tail.

Extreme value theory (EVT) is focused on accurately modeling the tail of an empirical distribution~\citep{Berlaint04BOOK}. In EVT, such tails are classified by a real number, called the tail index. In summary, there are two approaches to estimate the tail index: analyzing the empirical distribution of block minimums, and analyzing the empirical tail distribution. In any case, using models for tails requires the existence of an optimal threshold defining the starting point of the tail~\citep{Coles01BOOK}. In practice, one must verify that the sample size is large enough to accommodate a sub-sample of the tail of the distribution. In pre-analysis, we considered the Euclidean and DTW measures and confirmed that this property holds for all data sets and $w$. For each combination of measure, data, and $w$ we tried, the estimation of the optimal threshold provided an estimation of the tail index inside the confidence interval for the tail index obtained with the analysis of block minimums~\citep{DelCastillo14CSDA}. Thus, we found the considered data fulfilled the aforementioned requirement.

Obviously, the left tail distribution of the computed dissimilarities has a bounded range, since $d\geq 0$. That is called a short tail, and it corresponds to a negative value of the tail index. Therefore, the distributions considered as models for the lowest dissimilarities have to contain short tails. Since both tail distributions showed this behavior, we consider the simplest model to fit two-side short tails~\citep{Berlaint04BOOK}: the beta distribution. Besides the tails, we also observed that the behavior in the central part of the beta distribution was very similar to the behavior in the central part of most of the empirical distributions obtained for the considered cases. Thus, in addition to being a theoretically plausible model, the beta distribution was found to visually correspond to the empirical data.

\subsection{Model fit}
\label{sec:model_fit}

The beta distribution typically depends on two shape parameters, each of them corresponding to the tail index of each side. The extreme value for close-to-zero dissimilarity is zero, but in the case of the maximum, we have seen it depends on the original data set and $w$. 
Therefore, we consider the three-parameter beta distribution
\begin{equation}
	P(d) = \frac{1}{mB(\alpha,\beta)} \left(\frac{d}{m}\right)^{\alpha-1} \left(1-\frac{d}{m}\right)^{\beta-1} ,
	\label{eq:betapdf}
\end{equation}
where $\alpha,\beta>0$ are the so-called shape parameters, $m$ is a scale parameter, and $B(\alpha,\beta)$ is the beta function. Eq.~\ref{eq:betapdf} is defined for $0\leq d\leq m$. For values of $d$ outside this range, $P(d)=0$.

We start by fitting one beta distribution for each $w$. We do so by employing the maximum likelihood. Given $n$ normalized dissimilarities $d=d_1,\dots d_n$ computed from a uniform random sampling of all possible non-overlapping segments of length $w$ (Sec.~\ref{sec:mm_sampling}), we can calculate the log-likelihood
\begin{multline}
	\text{ln}(\mathcal{L}(\alpha,\beta,m|d)) = (\alpha-1) \sum_{i=1}^{n} \text{ln}(d_i) +\\
	+ (\beta-1) \sum_{i=1}^{n} \text{ln}(m-d_i) - n\text{ln}(B(\alpha,\beta)) - n(\alpha+\beta-1)\text{ln}(m) .
	\label{eq:likelihood}
\end{multline}
From here, we have to find the values of $\alpha$, $\beta$, and $m$ that maximize Eq.~\ref{eq:likelihood}. To do so, we choose a particle swarm optimizer~\citep{Poli07SI}.

Particle swarm optimization (PSO) is a well-known population-based stochastic approach for solving continuous and discrete optimization problems. PSO makes few or no assumptions about the problem being optimized, does not require it to be differentiable, can search very large spaces of candidate solutions, and can be applied to problems that are irregular, incomplete, noisy, dynamic, etc.~\citep[see][and references therein]{Poli07SI,Parsopoulos10BOOK}. We here use the canonical PSO algorithm~\citep{Poli07SI}, with 25~particles and a local best configuration, and run 300~iterations. Further details can be found in the provided code (Sec.~\ref{sec:mm_download}). The motivation for using PSO comes from our experience in optimization problems. However, we believe that more classical optimization algorithms would yield comparable, if not identical results. Essentially, any suitable optimization procedure available in typical scientific programming environments could do. The only constrains it needs to handle are $\alpha,\beta>0$ and $m>\text{max}(d)$. To facilitate the search, we additionally force $m<2.1\text{max}(d)$.
 
\begin{figure}[t]
	\centering
	\includegraphics{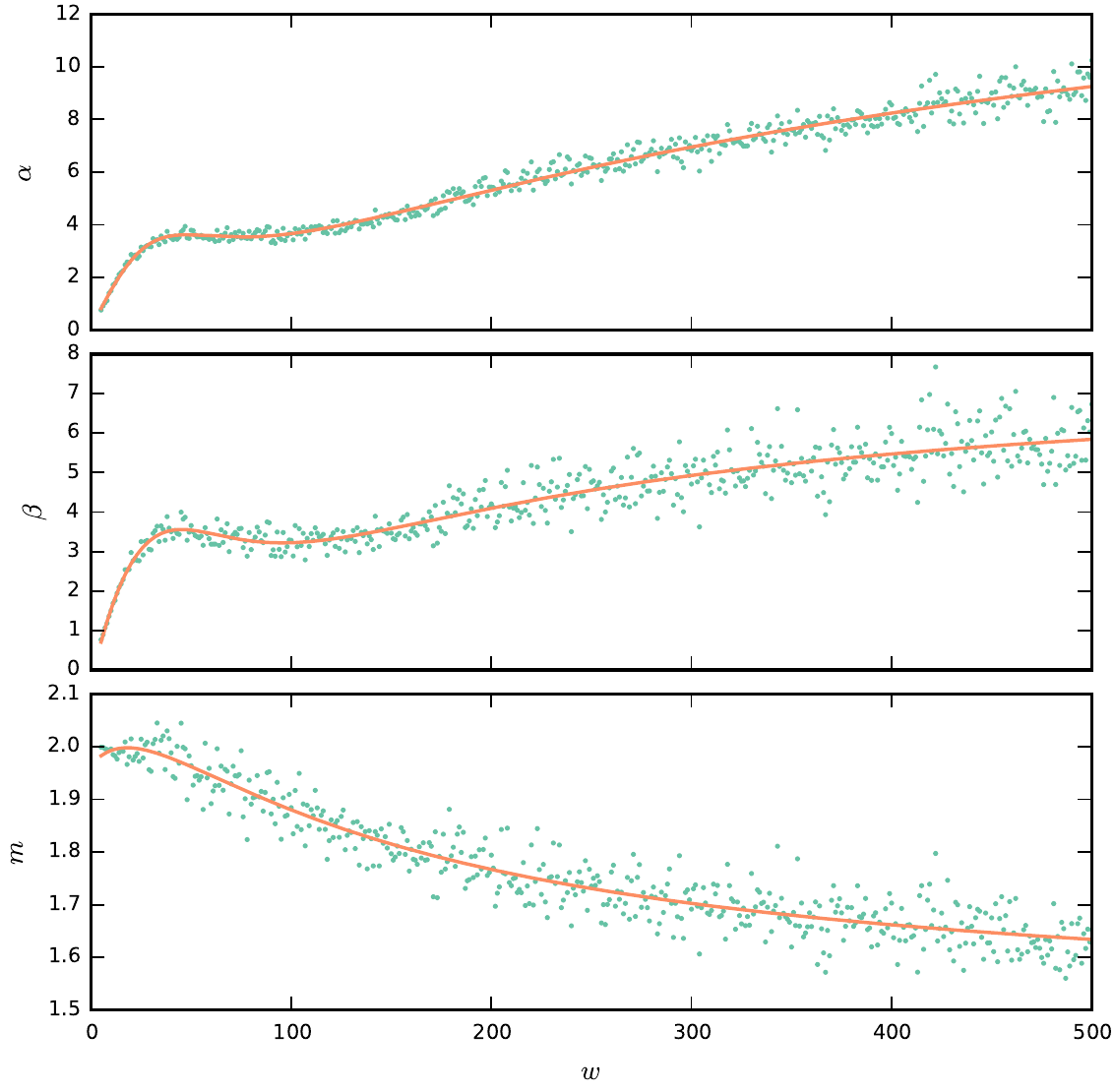}
	\caption{Example of the estimated parameters $\alpha$ (top), $\beta$ (middle), and $m$ (bottom) for each $w$. The fitted rational functions are also displayed. The samples come from using the cosine dissimilarity and the \textsc{Power} data set.}
	\label{fig:fit_series}
\end{figure}

If we repeat the previous procedure for all $w\in[w_{\text{min}},w_{\text{max}}]$, we end with three series of parameters: one for $\alpha$, one for $\beta$, and one for $m$ (Fig.~\ref{fig:fit_series}). This can represent a huge number of parameters for our model ($3\times(w_{\text{max}}-w_{\text{min}}+1)$). However, as we have seen in Sec.~\ref{sec:comparison_formal}, close distributions with $w_\Delta<40$ are rather similar, and this similarity increases as $w_\Delta$ decreases. Because of this, the estimated parameters exhibit a continuity in $w$ (Fig.~\ref{fig:fit_series}). We can exploit this continuity to fulfill two desirable objectives at the same time: reducing the number of parameters of our model, and removing some of the potential noise introduced in the sampling and/or the fitting procedure. This brings us to the next important step.

Given the three parameter series for $\alpha$, $\beta$, and $m$, we fit a curve to each of them by using rational functions, i.e.,~the ratio of two polynomial functions~\citep{Ghosh96BOOK}. A rational function model is a generalization of the polynomial model, as the former contains the latter as a subset. Rational function models provide several advantages over polynomial models while still having a moderately simple form~\citep{Ghosh96BOOK}. In particular, they are relatively easy to fit, take on an extremely wide range of shapes, and have very good interpolatory and extrapolatory properties. Thus, the three parameter beta distribution accounting for the full range of $w$ becomes
\begin{equation*}
	P_w(d) = \frac{1}{m_w B(\alpha_w,\beta_w)} \left(\frac{d}{m_w}\right)^{\alpha_w-1} \left(1-\frac{d}{m_w}\right)^{\beta_w-1} ,
\end{equation*}
where 
\begin{eqnarray}
	\alpha_w & = & Q_\alpha(w)/R_\alpha(w) , \label{eq:poly_a}\\
	\beta_w & = & Q_\beta(w)/R_\beta(w) ,\label{eq:poly_b}\\
	m_w & = & Q_m(w)/R_m(w) , \label{eq:poly_m}
\end{eqnarray}
and $Q_z$ and $R_z$ correspond to polynomials of degrees $u_z$ and $v_z$, respectively, such that
\begin{equation}
	Q_z(w) = \sum_{i=0}^{u_z} q_{z_i}w^i
	\label{eq:qz}
\end{equation}
and
\begin{equation}
	R_z(w) = 1 + \sum_{i=1}^{v_z} r_{z_i}w^i .
	\label{eq:rz}
\end{equation}

To fit the rational functions, we employ the default implementation of the Levenberg-Marquardt algorithm~\citep[LMA;][]{Gill78JNA} available in the Matlab's curve fitting toolbox\footnote{\url{http://www.mathworks.com/products/curvefitting}}. We recursively compute the fits for all pairwise combinations of $u_z=1,2,3$ and $v_z=0,1,2,3$, and take the one that yields the lowest Akaike information criterion~\citep{Burnham02BOOK}. For further details about this fitting procedure, we refer the interested reader to the provided code (Sec.~\ref{sec:mm_download}). The motivation for using the LMA is its improved robustness over the typical Gauss-Newton algorithm~\citep{Gill78JNA}. Again, as with the case of PSO, we believe that any other suitable curve fitting or optimization algorithm could be used with very similar or identical results.

The final model $P_w$ is parameterized by the rational functions $\alpha_w$, $\beta_w$, and $m_w$. Hence, it consists of $u_\alpha+1$, $v_\alpha$, $u_\beta+1$, $v_\beta$, $u_m+1$, and $v_m$ coefficients. From the values of $u_z$ and $v_z$ considered above, we see that the total number of model coefficients ranges from 6 ($3\times (2+0)$) to 21 ($3\times (4+3)$). A model with 6 to 21 coefficients can be considered a compact model given the size and complexity of the dissimilarity spaces we are dealing with (Sec.~\ref{sec:comparison}), which comprise $w_{\text{max}}-w_{\text{min}}+1$ different lengths or individual empirical distributions.

\subsection{Model usage}
\label{sec:model_usage}

As mentioned, our end goal is to `normalize' the dissimilarity space with respect to variations in $w$. To do so, we just need to compute $\alpha_w$, $\beta_w$, and $m_w$ following Eqs.~\ref{eq:poly_a}--\ref{eq:rz} and consider the CDF of the proposed model,
\begin{equation*}
	P_w(D\leq d) = \frac{ B \left( \frac{d}{m_w}; \alpha_w,\beta_w \right) }{ B\left(\alpha_w,\beta_w\right) } ,
\end{equation*}
where $B\left(x;\alpha,\beta\right)$ is the incomplete beta function, a generalization of the beta function. The incomplete beta function can be efficiently calculated using functions that are commonly included in spreadsheet or programming systems~\citep{Press07BOOK}. 

Because $P_w(D\leq d)$ is only defined for $0\leq d\leq m_w$, we propose the new dissimilarity measure
\begin{equation*}
	d' =
	\begin{cases}
		0 				& \text{if } d<0\\
		P_w(D\leq d)		& \text{if } 0\leq d\leq m_w \\
		1				& \text{otherwise}
	\end{cases}
\end{equation*}
for ranking and comparing motifs of different lengths under the same conditions.
The case of $d<0$ is impossible for most dissimilarity measures since typically $d\geq 0$. Moreover, if $d$ took negative values, we could always apply any suitable transformation to make it strictly positive (e.g.,~$e^d$). The case of $d>m_w$ might happen in practice, as our estimation of the maximum $d$ for each $w$ could be inaccurate or underestimating the true maximum (if this exists). However, this latter case is of little interest in motif discovery, as it corresponds to extremely dissimilar segment pairs. Thus, without compromising the accuracy of the task, we can tolerate some error and consider these motifs to form a tie in the last positions of the ranking ($d'=1$ for all of them).

The new dissimilarity measure $d'$ is a wrapper of $d$, and can be inserted in any motif discovery algorithm once $P_w$ has been estimated (offline or prior to the execution of the algorithm). Furthermore, $d'$ is easily interpretable, as it corresponds to the probability of seeing a dissimilarity equal to or smaller than $d$. This gives us an idea of the significance of the motif with respect to the dissimilarity space. 

\subsection{Model validation}
\label{sec:model_accuracy}

\begin{figure}[t]
	\centering
	\includegraphics{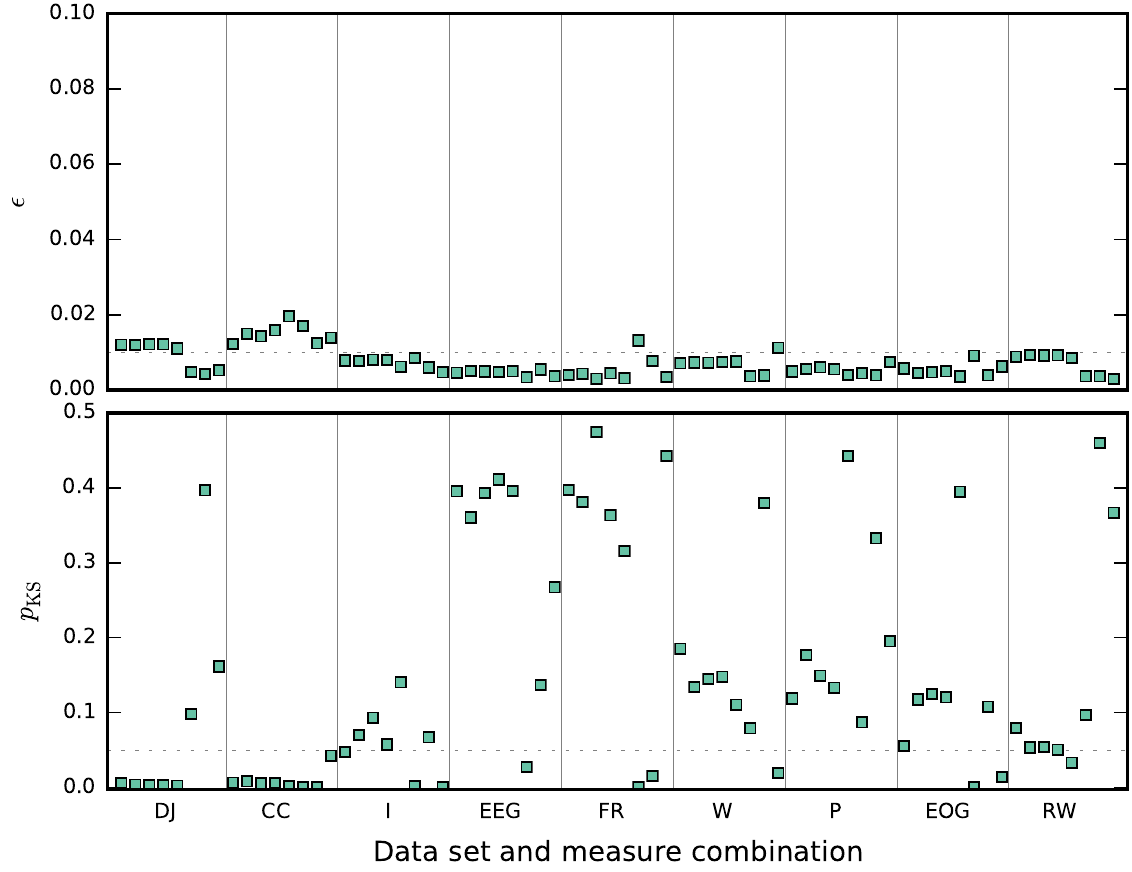}
	\caption{Model accuracy: median values for $\epsilon$ (top) and $p_{\text{KS}}$ (bottom) considering every combination of data set and dissimilarity measure.}
	\label{fig:model_epspval}
\end{figure}

\begin{figure}[t]
	\centering
	\includegraphics{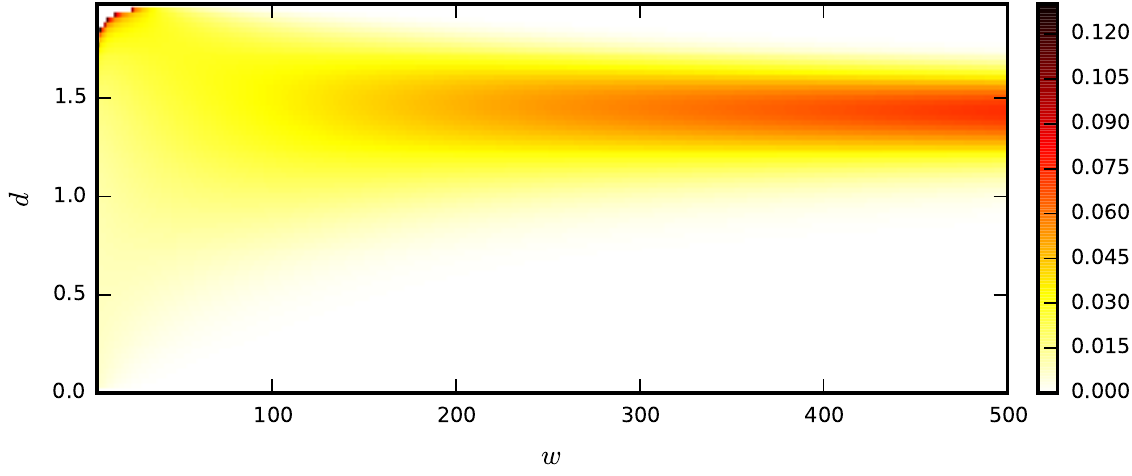}
	\caption{Fitted model for length-normalized Euclidean distances sampled from the \textsc{EEG} data set (compare with Fig.~\ref{fig:example_hist}).}
	\label{fig:model_hist}
\end{figure}

To measure the quality of the model fit $P_w$, we resort to the measures introduced in Sec.~\ref{sec:comparison_formal}: $\epsilon$, the global disagreement between empirical CDFs, and $p_{\text{KS}}$, the $p$-value of the KS test on the lowest quartile of the samples. The only difference is that here the $p_{\text{KS}}$ value is not the result of a two-sample test, but the result of a goodness of fit test for the plausibility of our proposed model given the available samples (we adapt the bootstrap generative procedure described by~\citet{Clauset09SIAMR} for power-law models to the current model). If we compute $\epsilon$ and $p_{\text{KS}}$ for all considered measures and data sets, we see that the fitted models generally provide a good agreement with the data (Fig.~\ref{fig:model_epspval}). In general, $\epsilon$ is never above 0.02 and rarely above 0.01. The $p_{\text{KS}}$ value is often above 0.05, what indicates that we cannot reject the null hypothesis of the tail samples coming from the fitted distribution tail. The \textsc{DowJones} and the \textsc{CarCount} data sets achieve relatively low $p_{\text{KS}}$ values, but $\epsilon$ is always below 0.02. The median and median absolute deviation for the aggregation of all combinations are $\epsilon=0.006\pm 0.002$ and $p_{\text{KS}}=0.10\pm 0.09$. Further details can be found in the online results (Sec.~\ref{sec:mm_download}). Overall, we can consider a reasonably good fit is reached for the majority of cases. We can visually confirm the agreement of our model and the empirical data by comparing the resultant PDFs against the empirical histograms obtained for each combination (compare, for instance, the obtained model in Fig.~\ref{fig:model_hist} with our motivating example of Fig.~\ref{fig:example_hist}). 

\section{Materials and methods}
\label{sec:mm}

\subsection{Time series data sets}
\label{sec:mm_data}

To demonstrate that our results are not biased with regard to the data source, we consider 9~different publicly-available time series of varying length, coming from distinct domains: 
(1) \textsc{DowJones} -- the daily closing values of the Dow Jones industrial average~\citep{Williamson12WEB}; 
(2) \textsc{CarCount} -- the number of cars measured for the Glendale on ramp for the 101~North freeway in Los Angeles, CA, USA~\citep{Ihler06KDD}; 
(3) \textsc{Insect} -- the electrical penetration graph of a beet leafhopper\footnote{\url{http://www.cs.ucr.edu/~mueen/MK}}~\citep{Mueen09SDM}; 
(4) \textsc{EEG} -- a one hour electroencephalogram from a single channel in a sleeping patient\footnote{\url{http://www.cs.ucr.edu/~mueen/OnlineMotif}}~\citep{Mueen09SDM}; 
(5) \textsc{FieldRecording} -- the spectral centroid of a field recording\footnote{\url{http://www.freesound.org/people/JeffWojo/sounds/121250}} (we used the mean of the stereo channels and the spectral centroid linear frequency plugin from Sonic Visualizer\footnote{\url{http://www.sonicvisualiser.org}}); 
(6) \textsc{Wind} -- the wind speed registered in the buoy of Rincon del San Jose\footnote{\url{http://lighthouse.tamucc.edu/pq}}, TX, USA. 
(7) \textsc{Power} -- the electric power consumption of an individual household\footnote{ \url{http://archive.ics.uci.edu/ml/datasets/Individual+household+electric+power+consumption }}~\citep{UCI13WEB}; 
(8) \textsc{EOG} -- an electrooculogram tracking the eye movements of a sleeping patient\footnote{\url{http://www.cs.ucr.edu/~mueen/DAME}}~\citep{Goldberger00CIRCULATION}; 
and (9) \textsc{RandomWalk} -- a random walk time series, artificially generated using $z_{i+1}=z_i+\eta$ and $z_1=0$, where $\eta$ is a Gaussian random number with zero mean and unit variance. 

\subsection{Dissimilarity measurement}
\label{sec:mm_dissim}

To demonstrate that our results are not biased with regard to the similarity measurement, we consider 8~different and commonly-used time series dissimilarity measures~\citep[see][and references therein]{Serra14KBS}: (1) Euc -- Euclidean distance normalized by $\sqrt{w}$; (2) sqEuc -- squared Euclidean distance normalized by $w$; (3) Corr -- Pearson's correlation, (4) Cos -- cosine dissimilarity, (5) DTW -- dynamic time warping with path-accumulated normalization weights and a $\pm$5\% corridor window; (6) EDR -- edit distance with real penalty normalized by the path length; (7) TWED -- time-warped edit distance normalized by the path length; and (8) MDL -- minimum description length as in~\cite{Rakthanmanon11ICDM}, with an added constant to force $d\geq 0$. All dissimilarities were computed between z-normalized non-overlapping time series segments.

\subsection{Motif sampling}
\label{sec:mm_sampling}

Given the formal definition of similarity-based time series motifs~\citep{Mueen09SDM}, to obtain possible motif candidates we just need to sample the motif space. In particular, we take $n=2000$ motif samples for each $w$ uniformly at random, and explicitly avoid trivial matches~\citep{Chiu03KDD}. That is, given a motif length $w$, we randomly generate the start of the segments that will form the motif, $i,j\in[1,N-w]$, $N$ being the time series length, such that $|i-j|>w$. If not stated otherwise, we consider $w_{\text{min}}=5$ and $w_{\text{max}}=500$, i.e.,~$w\in[5,500]$. 

\subsection{Further results, code, and data availability}
\label{sec:mm_download}

We will make available all raw results, code and data at our web page as soon as possible.


\section{Conclusion}
\label{sec:conclusion}

The main contribution of the present work is to show that time series motif dissimilarities of different lengths are not directly comparable, and thus cannot be ranked. Through both motivating examples and formal quantitative analysis, we have shown (1) that length-normalized motif dissimilarities have non-linear dependencies with the motif length, (2) that these dependencies change with the data set and the dissimilarity measure, and (3) that they particularly affect the lowest dissimilarities, which are precisely the focus of interest in similarity-based motif discovery. Another contribution of the present work is a solution to tackle the aforementioned problems. This consists of a compact model of the dissimilarity space that allows comparing motifs of different lengths and assessing their significance with respect to the overall dissimilarity distribution. Such model is motivated by extreme value theory, and is based on a three-parameter beta distribution. We propose a procedure to fit those three parameters while taking into account the local continuity and the non-linearity of the motif dissimilarity space.

In this work, we have not explicitly dealt with motif pairs consisting of segments of different length. Instead, we have assumed the same length for the pair of segments forming a motif pair. This assumption is well motivated, as practically all existing motif discovery algorithms operate under such constraint~\cite[e.g.,][]{Lin02WTDM,Chiu03KDD,Tanaka05ML,Mueen09SDM,Castro11SDM,Mueen13ICDM,Yingcharxxx13ICDM}. It is also motivated for the case where we are interested in pairs of segments of different length, as the most common way to compute the dissimilarity between such segments is by re-sampling them to have the same length. That is extensively used for Euclidean distance or correlation~\citep{Yankov07KDD}. For measures explicitly handling segments of different length, this is also one of the most recommended practices. For instance, it has been shown that a brute-force up-sampling to the largest segment length yields equivalent or slightly better results for classification tasks using DTW~\citep{Ratanamahatana04WMTSD}.


It is difficult to assess the potential impact of the present findings in other contexts. However, we have the impression that a similar phenomenon could happen when comparing feature vectors or quantitative descriptions of different sizes, even if these are not time series or segments. It would be very interesting to analyze what happens with clustering or classification tasks with variable-length instances, and in particular with clustering or classification approaches based on dissimilarity measurements. The scarce literature on the topic we have found typically relies domain-specific knowledge~\cite[e.g.,][]{McHardy07NATURE} or makes a number of assumptions on the nature of the data~\cite[e.g.,][]{Porikli04TR}. The model and the methodology proposed here are domain-agnostic and make very few assumptions. Thus, we believe they could be good candidates to be considered in situations where variable-length instance similarities need to be compared.

\section*{Acknowledgments}

We would like to thank all the people who contributed the data sets used in this study. This research has been funded by Generalitat de Catalunya and the Spanish Government: 2009-SGR-1434 (JS, JLA), 2014-SGR-1307 (AC, IS), FIS2012-31324 (AC), MTM2012-31118 (IS), and TIN2012-38450-C03-03 (JS, JLA).

\section*{References}

\bibliography{bibjserra}

\end{document}